\documentclass{bmvc2k}

\usepackage{makecell}
\usepackage{svg}
\usepackage[printonlyused,withpage,nolist]{acronym}
\usepackage{cleveref}
\usepackage{booktabs}
\usepackage{multirow}
\usepackage{lipsum} 
\definecolor{uhhstone}{RGB}{59,81,91}
\usepackage{url} 

\title{TRUDI and TITUS: \\ A Multi-Perspective Dataset and A Three-Stage Recognition System for Transportation Unit Identification}

\addauthor{Emre Gülsoylu}{emre.guelsoylu@uni-hamburg.de}{1}
\addauthor{André Kelm}{andre.kelm@uni-hamburg.de}{1}
\addauthor{Lennart Bengtson}{lennart.bengtson@uni-hamburg.de}{2}
\addauthor{Matthias Hirsch}{matthias.hirsch@uni-hamburg.de}{1}
\addauthor{Christian Wilms}{christian.wilms@uni-hamburg.de}{1}
\addauthor{Tim Rolff}{tim.rolff@uni-hamburg.de}{3}
\addauthor{Janick Edinger}{janick.edinger@uni-hamburg.de}{2}
\addauthor{Simone Frintrop}{simone.frintrop@uni-hamburg.de}{1}
\addinstitution{
 Computer Vision Group\\
 University of Hamburg\\
 Hamburg, Germany
}
\addinstitution{
 Distributed Operating Systems Group\\
 University of Hamburg\\
 Hamburg, Germany
}
\addinstitution{
 Human-Computer Interaction Group\\
 University of Hamburg\\
 Hamburg, Germany
}

\runninghead{Gülsoylu et al.}{TRUDI and TITUS}

\begin{document}
\begin{acronym}
    \acro{tu}[TU]{transportation unit}
    \acro{uav}[UAV]{unmanned aerial vehicle}
    \acro{dl}[DL]{deep learning}
    \acro{ml}[ML]{machine learning}
    \acro{iou}[IoU]{intersection over union}
    \acro{map}[mAP]{mean average precision}    
    \acro{ap}[AP]{average precision}
    \acro{ar}[AR]{average recall}
    \acro{hmean}[Hmean]{harmonic mean}
    \acro{id}[ID]{identifier}
    \acro{rs}[RS]{reach stacker}
    \acro{titus}[TITUS]{Three-stage Identification of Transportation UnitS}
    \acro{trudi}[TRUDI]{TRansportation Unit Detection and Identification}
    \acro{ocr}[OCR]{Optical Character Recognition}
    \acro{dslr}[DSLR]{Digital single-lens reflex camera}
    \acro{3c}[3C]{Three Classes}
    \acro{1c}[1C]{One Class}  
\end{acronym}

\maketitle

\begin{abstract}
\footnotesize{
Identifying transportation units (TUs) is essential for improving the efficiency of port logistics. However, progress in this field has been hindered by the lack of publicly available benchmark datasets that capture the diversity and dynamics of real-world port environments. To address this gap, we present the TRUDI dataset—a comprehensive collection comprising 35,034 annotated instances across five categories: container, tank container, trailer, ID text, and logo. The images were captured at operational ports using both ground-based and aerial cameras, under a wide variety of lighting and weather conditions.
For the identification of TUs—which involves reading the 11-digit alphanumeric ID typically painted on each unit—we introduce TITUS, a dedicated pipeline that operates in three stages: (1) segmenting the TU instances, (2) detecting the location of the ID text, and (3) recognising and validating the extracted ID. Unlike alternative systems, which often require similar scenes, specific camera angles or gate setups, our evaluation demonstrates that TITUS reliably identifies TUs from a range of camera perspectives and in varying lighting and weather conditions. By making the TRUDI dataset publicly available, we provide a robust benchmark that enables the development and comparison of new approaches. This contribution supports digital transformation efforts in multipurpose ports and helps to increase the efficiency of entire logistics chains.
}
\end{abstract} 


\section{Introduction} \vspace{-0.5em}
\label{sec:intro}
Multipurpose terminals in ports need to accommodate various modes of transportation and volatile cargo volumes with limited available space. The resulting dynamic environments created by constant modifications in storage configurations and operational processes make the use of fixed infrastructure for monitoring terminals impractical. This leads to increased manual labour for inventory keeping and inadequate traceability of \acp{tu}, such as containers and cranable semi-trailers. Efficient, digital terminal monitoring thus requires the reliable identification of individual \acp{tu}~\cite{shetty2012optical, mcfarlane2003impact}. 

\acp{tu} have standardised dimensions and feature unique markings as defined in ISO6346~\cite{iso6346freight}. These alphanumeric ID codes ensure unambiguous visual identification and are usually painted on the top or sides of each \ac{tu}. They consists of a four-letter owner code, a six-digit serial number, and a single check digit. Despite advancements in automation systems for identification technologies like \ac{ocr} gates~\cite{zhao2024practical} and RFID tags~\cite{shi2011rfid} on the \acp{tu}, these approaches often fall short in adaptability, especially in multipurpose terminals with seasonal operational variability. Existing solutions for \ac{tu} identification typically rely on character detection from a specific target area followed by character recognition~\cite{zhao2024practical, hoa2023build}. 
Other solutions relax the assumption of a predefined target area by adding another step to detect the ID first~\cite{yu2024two, yang2023lightweight}. However, these methods are still constrained to fixed camera placements and struggle with changing perspectives, particularly when using mobile cameras mounted on \acp{uav} or \acp{rs}~\cite{teegen2024drone}. Moreover, existing methods are mostly evaluated on proprietary datasets often consisting of images taken in one single port which limits comparability. There are currently no publicly available datasets for \ac{tu} identification~\cite{liu2025lightweight}. 


\begin{sloppypar}
We introduce the \ac{trudi} dataset comprising images captured from both aerial and ground perspectives under different lighting and weather conditions. This dataset addresses the comparability issue and supports the progress towards more adaptable port monitoring operations utilising moving cameras. Moreover, we propose a flexible \ac{tu} identification method which is suitable for use with both aerial and ground-based imagery and, thus, does not rely on fixed infrastructure. We introduce and employ the \ac{titus} pipeline that consists of segmentation of \ac{tu} instances, ISO6346 compliant ID text detection, and text recognition. The use of an instance segmentation stage prior to text detection and text recognition enables associating \acp{tu} with their IDs reducing the search space for text detection. Additionally, the association of the segmented instances and ID codes supports the localisation of \acp{tu} inside terminals using mobile cameras. This facilitates down-stream applications such as the creation of a digital twin for the detailed analysis of operational processes. With the release of \ac{trudi} and the introduction of \ac{titus}, we aim to provide researchers and practitioners with valuable resources for developing and evaluating new methods on multiperspective and robust identification of \ac{tu}.
\end{sloppypar}


\begin{sloppypar}
In summary, our contributions are: (1) a new and publicly available dataset\footnote{\href{https://github.com/egulsoylu/trudi}{https://github.com/egulsoylu/trudi}}, \ac{trudi}, for \ac{tu} identification from aerial and ground perspectives, (2) a novel three-stage pipeline, \ac{titus}, and the (3) detailed evaluation of the proposed pipeline on the \ac{trudi} dataset.
\end{sloppypar}


\section{Related Work} \vspace{-0.5em}
Existing literature mostly focuses on automatic container code recognition using fixed cameras~\cite{liu2025lightweight, yao2025advanced, diaz2024wagon}, thereby excluding the use of mobile cameras, including ground-based handheld devices and aerial platforms such as \acp{uav}. While the use of UAVs, in particular, enables more scalable and flexible image acquisition, it also introduces new challenges, such as the detection of small text areas within complex backgrounds. These challenges can reduce performance in identifying \acp{tu}~\cite{teegen2024drone}. Furthermore, the identification of intermodal loading units, such as cranable semi-trailers, remains largely overlooked in the literature~\cite{zhang2021vertical}.

Early approaches for this task often rely on digital image processing techniques~\cite{lui1990neural, lee1999automatic, pan2004robust}. These conventional methods are still employed alongside \ac{dl} methods to form hybrid solutions, allowing for effective task-specific feature engineering. Nguyen et al.~\cite{nguyen2023digital} employ both conventional computer vision and \ac{ml} techniques including histogram of oriented gradients and support vector machines in a pre-processing-intensive method for text detection and recognition. While their approach has a robust pre-processing stage, the lack of comparative evaluation with \ac{dl} models presents limitations, particularly regarding the adaptability and accuracy of their system in uncontrolled environments. Additionally, Hsu et al.~\cite{hsu2023automatic} employ YOLOv4 for the initial detection phase and use Tesseract OCR~\cite{smith2007overview} for text recognition, integrating histogram equalisation and morphological operations in the pre-processing stage. Although this method benefits from powerful OCR capabilities, its performance and robustness in varying environmental conditions remain unclear due to limited dataset diversity. Another hybrid method~\cite{li2022towards}, an end-to-end recognition system, applies edge detection and component analysis to classify characters with support vector machines. This approach demonstrates an alternative to \ac{dl}-based methodologies yet remains susceptible to variations in lighting and \ac{tu} condition.

Given that \ac{ml}-based solutions have demonstrated automatic feature learning and extraction from a given dataset, researchers have increasingly focused on these methods in recent years. An approach by Zhao et al.~\cite{zhao2024practical} introduces the Practical Unified Network (PUN), designed to localise and recognise arbitrary-oriented container codes, integrating detection and classification within a single framework. This model uses a ResNet18~\cite{he2016deep} backbone and demonstrates superior performance over traditional CNN- and transformer-based methods such as EAST~\cite{zhou2017east}, DETR~\cite{zhu2020deformable} and ABCv1~\cite{liu2020abcnet}. Its end-to-end design provides an efficient solution for static camera settings. However, it is less efficient for mobile cameras which can operate in a perception-action loop~\cite{gaussier1998perception} to iteratively select better perspectives for image capturing and, thus, reduce the number of unsuccessful text ID recognition attempts.
Yang et al.~\cite{yang2023lightweight} focus on real-time processing with a lightweight model based on multi-reuse feature fusion and a multi-branch structure merger. For this, they optimise detection with MobileOne blocks~\cite{vasu2023mobileone} and recognition using MobileNetV3-small~\cite{koonce2021mobilenetv3}. Even though they demonstrate accuracy improvements over YOLOv5~\cite{glenn_jocher_2021_5563715} their methodology does not fully address the challenges involved in the mobile camera-based applications as it treats text recognition as character detection. Character-level detection, however, is not suitable for scene-text detection as the text is scattered in the scene image, and there is no prior information about their location~\cite{lin2020review}. While the system is capable of high processing speeds, its applicability in real-world contexts may be constrained, as  however, as the evaluation relies on a comparatively small and non-diverse dataset. Li et al.~\cite{liu2025lightweight} tackle \ac{tu} identification as a character detection problem and introduce ACCR-YOLOv7 incorporating a feature extraction module called G-ELAN and an improved Efficient Spatial Pyramid Pooling Module. This model reduces computational complexity by replacing YOLOv7's ordinary 3x3 convolution with GSconv in the neck. 

In summary, while significant progress has been made for the \acfp{tu} identification task, developing robust and adaptable solutions with consistent performance across diverse operational scenarios, environments and perspectives remains a challenge. Current systems are unusable with vehicle-mounted cameras as they require \acp{tu} to be placed in a predefined area. Moreover, the lack of benchmark datasets hinders the comparability of the proposed methods. This leads to very high accuracies reported for methods using less complex datasets~\cite{liu2025lightweight} and low numbers for methods evaluated on complex datasets~\cite{teegen2024drone}. Therefore, \ac{trudi} can serve as a benchmark to enhance the reliability and comparability of existing systems and facilitate further innovation and progress in \ac{tu} identification.


\section{TRUDI Dataset} 
\begin{figure*}
    \centering
    \includegraphics[width=1\linewidth]{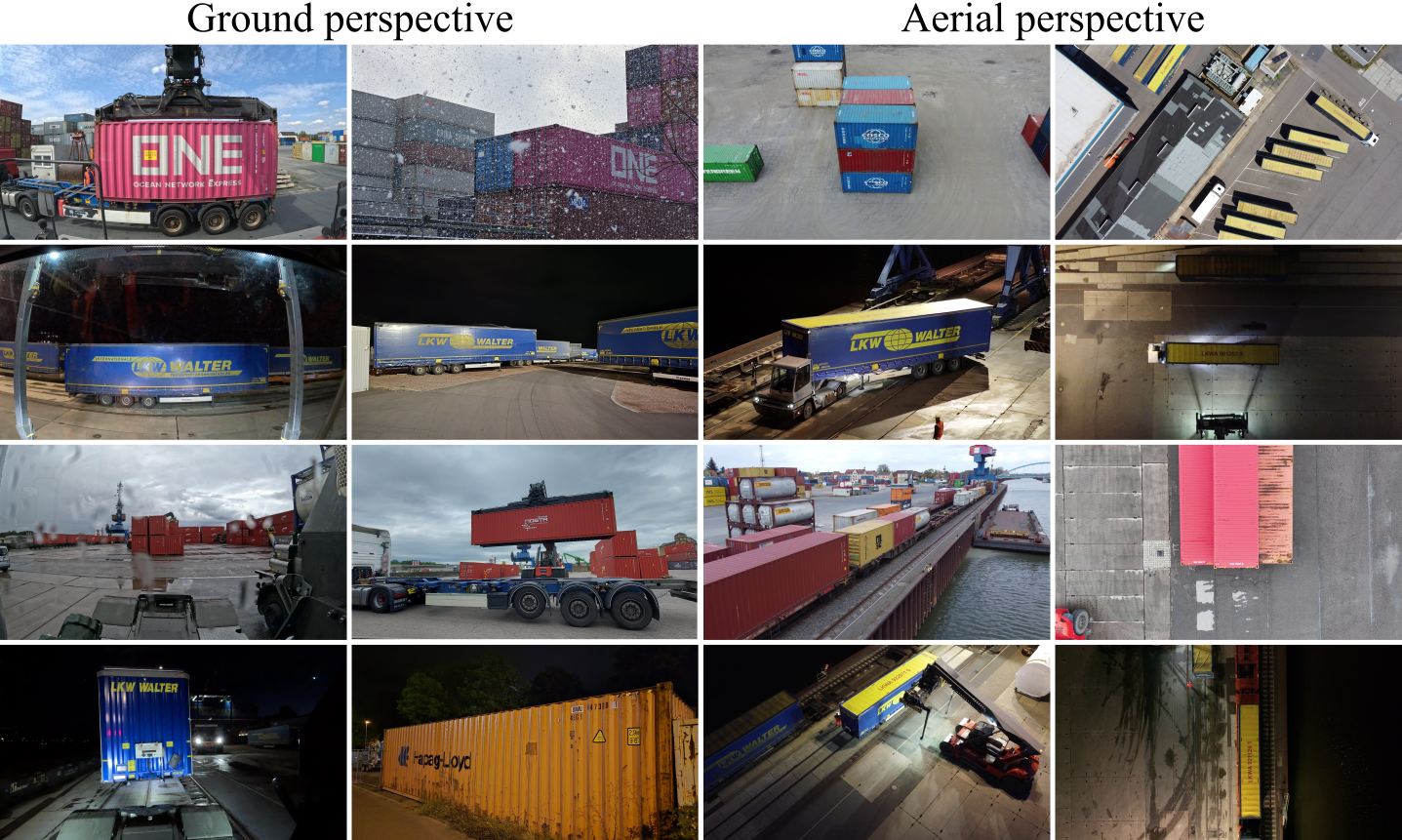}
    \caption{Sample images from the \ac{trudi} dataset showing ground and aerial perspectives.}
    \label{fig:trudi_sample}
\end{figure*}

\vspace{-0.5em}
To address the lack of publicly available datasets suitable for \ac{tu} identification using images from mobile cameras, we collected and annotated a comprehensive and multifaceted dataset featuring images captured from both aerial and ground-based perspectives. As shown in \cref{tab:dataset_statistics}, the dataset comprises 35,034 labelled instances of \acp{tu} and their markings, with an average of approximately 48 instances per image. 17,604 instances were collected from ground perspective through various devices, including smartphones, digital single-lens reflex cameras (DSLR), and camera-equipped vehicles such as terminal trucks and \acp{rs}. These images were captured during the vehicle's active use in port operations. The remaining 17,430 instances were captured by \acp{uav} using models like the DJI Mavic Pro 3, DJI Mini 2, and DJI Air 3\footnote{\href{https://www.dji.com}{dji.com}}. Sample images from the \ac{trudi} dataset are shown in \Cref{fig:trudi_sample} from both ground and aerial perspective.

The images in \ac{trudi} cover a wide range of perspectives, zoom levels, resolutions, and image qualities, which provides a diverse dataset for object detection, instance segmentation, logo detection, text detection, text recognition and text spotting. This diversity enables trained models to handle real-world scenarios by improving their robustness and generalisation capabilities. The objects are labelled as masks belonging to one of the five classes during the annotation process. Three of these classes represent common \ac{tu} types (\textit{container}, \textit{trailer}, \textit{tank\_container}), while the \ac{tu} markings are represented by two classes (\textit{id\_text}, \textit{logo}). Not all \ac{tu} masks have associated markings due to occlusions or large distances that make IDs undetectable and illegible.

The number of instances is shown in \cref{tab:class_wise_statistics}. All annotators and reviewers who check the quality of the annotations have been provided with comprehensive guidelines to avoid any inconsistencies in annotations that negatively affect the reliability of models~\cite{liu2022inconsistent}. These guidelines outline the annotation process and provide clear definitions for classes of interest and edge cases such as occlusions or damaged and illegible markings\footnote{Further details about the dataset can be found in the supplementary material.}.





\begin{table*}
    \centering
    \caption{Dataset statistics per perspective and subset. The combined set contains the images from both aerial and ground perspectives.}
    {\footnotesize
    \begin{tabular}{@{}ll|rrrr@{}}
        \toprule
        Perspective & Subset      & \# Images & \# Instances & Avg. Instances & Median Instances \\ 
        \midrule
                    & Training    & 231        & 9864           & 42.70             & 24.0               \\
        Aerial      & Validation  & 77        & 3869           & 50.25             & 24.0               \\
                    & Test        & 75        & 3697           & 49.29             & 23.0               \\ 
                    \midrule              
                    & Training    & 210        & 9942           & 47.34             & 15.0               \\
        Ground      & Validation  & 70        & 3663           & 52.33             & 17.0               \\
                    & Test        & 70        & 3999           & 57.13             & 21.5               \\ 
                    \midrule      
                    & Training    & 441        & 19806           & 44.91             & 20.0               \\
        Combined    & Validation  & 147        & 7532           & 51.24             & 23.0               \\
                    & Test        & 145        & 7696           & 53.08             & 22.0               \\ 
                    \midrule
        Total       &             & 733 & 35034 & 47.80 & 21.0 \\ 
                    \bottomrule
    \end{tabular}
    }
    \label{tab:dataset_statistics}
\end{table*}


\begin{table*}
    \centering
    \caption{Class-wise instance count and size categories following COCO style~\cite{lin2014microsoft} (px).}
    {\footnotesize
    \begin{tabular}{@{}lrrrr@{}}
        \toprule
       Category        & \# Instances & Small ($area<32^{2}$) & Medium & Large ($96^{2}<area$) \\
        \midrule
        Container       & 11109 & 913   & 4868   & 5328  \\
        Tank Container  & 808   & 65    & 351    & 392   \\
        Trailer         & 2780  & 67    & 455    & 2258  \\
        ID Text         & 14009 & 9245  & 3433   & 1331  \\
        Logo            & 6328  & 2096  & 2361   & 1871  \\
        \bottomrule
    \end{tabular}
    }
    \label{tab:class_wise_statistics}
\end{table*}


To ensure environmental and temporal diversity, the images in \ac{trudi} were collected over an 18-month period across different countries, at various times of the day, including daytimes, dusk, and night, as well as during different seasons. Furthermore, the images were taken under various weather conditions, ranging from sunny and partly cloudy to overcast, rainy, and snowy. \acp{uav} were operated at various altitudes ranging from 3 to 120\,m, with an average flight altitude of approximately 30\,m, to represent different scenarios suitable for \acp{tu} identification missions. We excluded any frames showing the same objects from the identical viewpoint to avoid data redundancy. 

Randomly dividing a dataset could lead to a trivial test set or result in certain types of images being included only in one of the subsets. Therefore, we divided our \ac{trudi} dataset into three subsets (60\,\% train, 20\,\% validation, 20\,\% test) while ensuring that the subsets have a similar distribution.
The dataset was divided by binning the images based on their brightness, sharpness, and contrast, using a uniform bin range. We chose these features, as they are frequently used in image enhancement to assess image quality~\cite{nimkar2013contrast}. We then combined these bins into a single category for each image. This categorisation enabled us to stratify the subsets effectively.

\section{TITUS} 
\begin{figure}
    \centering
    \includegraphics[width=1\textwidth]{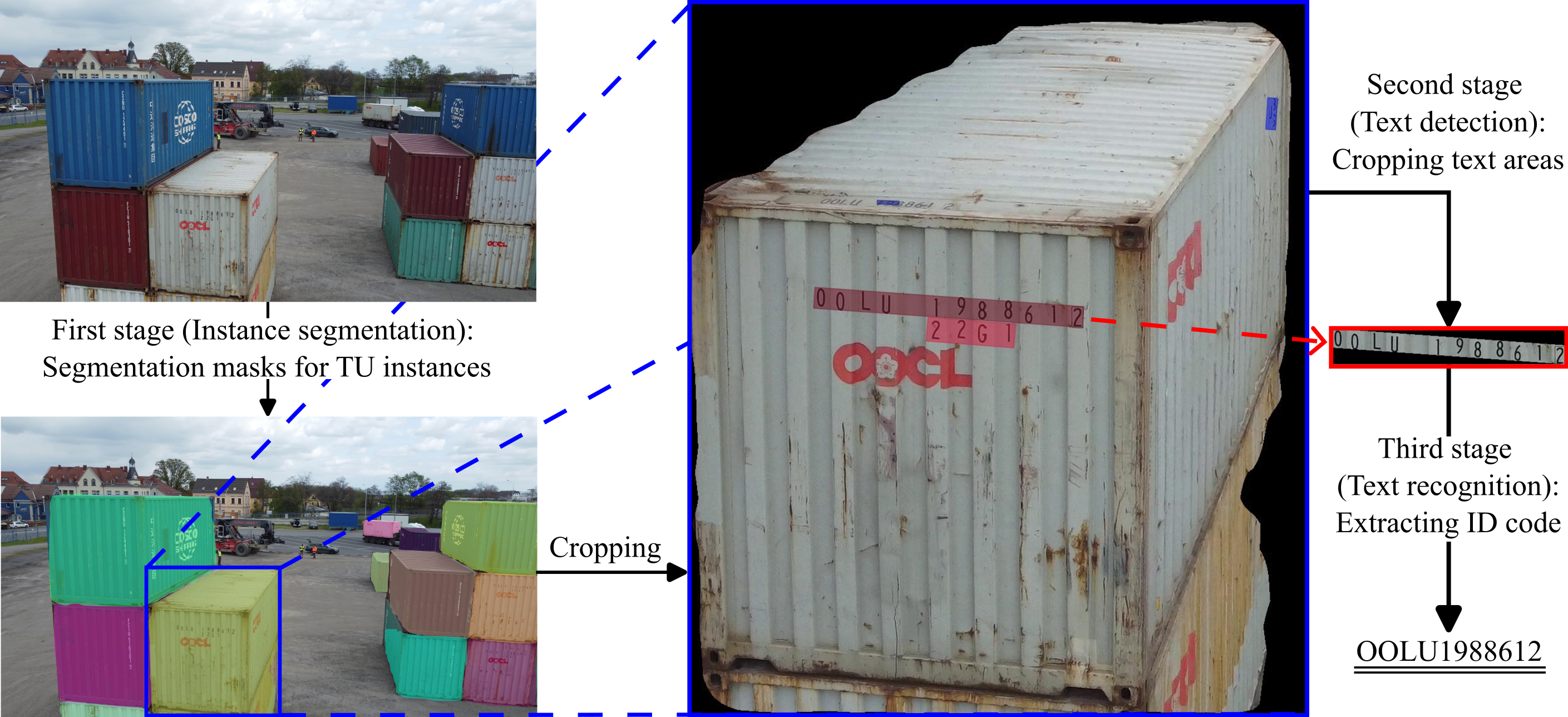}
    \caption{Overview of the \ac{titus} pipeline. \acp{tu} instances are segmented, then the ISO6436 compliant ID texts are detected, finally, a text recognition model extracts detected IDs.}
    \label{fig:overview}
\end{figure}

\vspace{-0.5em}
Our novel system involves three stages: (1) segmenting \ac{tu} instances (containers, tank containers, and trailers), (2) detecting their ID text area, (3) extracting the ID code from detected text areas and associating the extracted ID with the corresponding \ac{tu} instance. \Cref{fig:overview} illustrates this pipeline. 

Segmenting \acp{tu} instances as a prior step benefits the text detection since it reduces the search space to cropped regions. This increases inference speed and enhances text detection quality by minimising the number of distractors, compared to detecting ID text in the entire image.  The resulting cutouts effectively isolate the \acp{tu}, avoiding the inclusion of background areas that can still be present in bounding boxes. As a result, IDs from \acp{tu} in the background are less likely to interfere with associating each \ac{tu} to its own ID code, which is especially important in densely packed storage scenarios.

We selected the models based on their performance on the \ac{trudi} dataset. The first stage employs Mask R-CNN~\cite{lyu2022rtmdet} to segment \acp{tu} in images. The model is pre-trained on COCO~\cite{lin2014microsoft} and fine-tuned on the \ac{trudi} dataset only with the \textit{container}, \textit{tank\_container}, and \textit{trailer} classes. This stage outputs cropped masks of the \ac{tu} instances for the text detection stage. The second stage uses the DBNet++~\cite{liao2022real} model that was pre-trained on SynthText~\cite{gupta2016synthetic} and fine-tuned on \ac{trudi} for text detection within the \ac{tu} masks, addressing challenges such as varying orientations and environmental conditions including bad condition of \ac{tu} markings. In the final stage, detected text areas are cropped and fed into the RobustScanner~\cite{pan2004robust}, a text recognition model pre-trained on the ICDAR15 dataset~\cite{karatzas2015icdar} and fine-tuned on \ac{trudi}. The model recognises text and extracts the ID code. The output is then verified for ISO6436 compliance and associated with the \ac{tu}. If the text does not comply, its location is noted for further inspection. The pipeline outputs a file that associates \acp{tu} and their respective markings. This association is crucial for port monitoring as it enables real-world localisation of each \ac{tu} via georeferenced images.

\section{Experiments and Results} \vspace{-0.5em}
We conducted experiments using \ac{trudi} and its subsets (aerial, ground, and combined perspectives) to evaluate the three stages of \ac{titus} and establish a baseline. Instance segmentation models were fine-tuned on full images, while text detection and recognition models were trained on cropped TU instances and text areas, respectively.


\begin{sloppypar}
\vspace{2mm}
\noindent\textbf{Instance Segmentation:}
The initial stage aims to provide a mask for each \ac{tu} individually, using an instance segmentation model. For this stage, we fine-tuned two models that were pre-trained on the COCO~\cite{lin2014microsoft} dataset: Mask R-CNN with a Swin Transformer~\cite{he2017mask, liu2021swin} backbone and RTMDet~\cite{lyu2022rtmdet}. We fine-tuned the instance segmentation models in two settings: \ac{3c} setting involved fine-tuning the models to differentiate between three \ac{tu} classes (\texttt{container}, \texttt{tank\_container}, and \texttt{trailer}). The second setting, \ac{1c}, consolidated all three types of \ac{tu} classes into a single class as the ID code associated with each unit provides sufficient information to determine its type, thereby simplifying the classification task.    
\end{sloppypar}

\begin{table}
    \centering
    \caption{Results for stage 1: instance segmentation. "-C" denotes fine-tuning on Combined dataset. "3C" (Three Classes) indicates that the models are trained to distinguish between \texttt{container}, \texttt{tank\_container} and \texttt{trailer} classes, while "1C" (One Class) covers all three types of \acp{tu} in one class.}
    {\footnotesize
    \begin{tabular}{@{}llccc@{}}
        \toprule
        \makecell{Perspective} & \makecell{Model} & \makecell{AP@0.50:0.95} & \makecell{AP@0.50} & \makecell{AR@0.50:0.95} \\
        \midrule
        \multirow{5}{*}{Aerial} 
        & RTMDet (1C) & 0.22 & 0.29 & 0.44 \\
        & Mask R-CNN-Swin (1C) & \textbf{0.43} & \textbf{0.62} & \textbf{0.48} \\
        & RTMDet (3C) & 0.34 & 0.44 & 0.42 \\
        & Mask R-CNN-Swin (3C) & 0.38 & 0.56 & 0.44 \\
        & Mask R-CNN-Swin (3C)-C & 0.42 & \textbf{0.62} & 0.47 \\
        \cmidrule(lr){1-5}
        \multirow{5}{*}{Ground} & RTMDet (1C) & 0.27 & 0.43 & 0.45 \\        
        & Mask R-CNN-Swin (1C) & 0.40 & 0.62 & 0.48 \\
        & RTMDet (3C) & 0.24 & 0.37 & 0.38 \\
        & Mask R-CNN-Swin (3C) & 0.40 & 0.61 & 0.47 \\
        & Mask R-CNN-Swin (3C)-C & \textbf{0.44} & \textbf{0.67} & \textbf{0.50} \\
        \cmidrule(lr){1-5}
        \multirow{4}{*}{Combined} & RTMDet (1C) & 0.35 & 0.52 & \textbf{0.50} \\
        & Mask R-CNN-Swin (1C) & 0.43 & 0.64 & 0.49 \\
        & RTMDet (3C) & 0.41 & 0.57 & 0.49 \\
        & Mask R-CNN-Swin (3C) & \textbf{0.44} & \textbf{0.65} & \textbf{0.50} \\
        \bottomrule
    \end{tabular}
    }
    \label{tab:instance_segmentation_results}
\end{table}



\cref{tab:instance_segmentation_results} shows the results of the instance segmentation model, evaluated using \ac{ap} and \ac{ar} across various \ac{iou} thresholds. Mask R-CNN consistently outperforms RTMDet across all settings, especially for average precision, making it the preferred model for this stage in \ac{titus}. For both aerial and ground perspective, training with the one class setting achieves higher precision and recall. However, when the perspectives are combined, both settings produce comparable results. The models benefit from being exposed to diverse viewpoints during training as the combined perspective results in the highest overall scores. Fine-tuning on both aerial and ground datasets and testing on the ground perspective results in superior performance compared to fine-tuning only on the ground perspective dataset. This suggests that the ground perspective benefits from the inclusion of aerial perspective images during the fine-tuning process. Conversely, fine-tuning on the combined dataset and testing on the aerial dataset shows that the aerial perspective does not gain benefits from the ground perspective images.

\vspace{2mm}
\noindent\textbf{Text Detection:} For the text detection stage, we fine-tuned three text detection models which were pre-trained on SynthText~\cite{gupta2016synthetic}: PANet~\cite{liu2018path}, a model originally designed for instance segmentation, DBNet~\cite{liao2020real}, a prominent real-time scene-text detection model, DBNet++~\cite{liao2022real}, an improved version of DBNet with better feature fusion and differentiable binarisation. For fine-tuning this stage, each \ac{tu} mask was cropped based on the ground truth annotations. These crops were then fed into the model as input for fine-tuning.

\begin{table}
    \centering
    \caption{Results for stage 2: text detection (@0.5 IoU). "-C" indicates that the model is fine-tuned on Combined dataset but tested on either aerial or ground perspective.}
    {\footnotesize
    \begin{tabular}{@{}llccc@{}}
        \toprule
        \makecell{Perspective} & \makecell{Model} & \makecell{Recall} & \makecell{Precision} & \makecell{Hmean} \\
        \midrule
        \multirow{4}{*}{Aerial} & DBNet++ & 0.77 & 0.61 & 0.68 \\
        & DBNet++-C & \textbf{0.79} & \textbf{0.68} & \textbf{0.73} \\
        & DBNet & 0.29 & 0.30 & 0.29 \\
        & PANet & 0.44 & 0.14 & 0.21 \\
        \cmidrule(lr){1-5}
        \multirow{4}{*}{Ground} & DBNet++ & 0.73 & 0.63 & 0.68 \\
        & DBNet++-C & \textbf{0.79} & \textbf{0.69} & \textbf{0.74} \\
        & DBNet & 0.18 & 0.44 & 0.25 \\
        & PANet & 0.48 & 0.05 & 0.09 \\
        \cmidrule(lr){1-5}
        \multirow{3}{*}{Combined} & DBNet++ & \textbf{0.79} & \textbf{0.69} & \textbf{0.74} \\
        & DBNet & 0.43 & 0.55 & 0.48 \\
        & PANet & 0.57 & 0.16 & 0.25 \\
        \bottomrule
    \end{tabular}
    }
    \label{tab:text_detection_results}
\end{table}



\cref{tab:text_detection_results} presents results for text detection with the evaluation metrics recall, precision and \ac{hmean} for detection models. DBNet++ achieves the best performance across all three settings, which makes it the preferred text detection model for the second stage of \ac{titus}. Training on the combined dataset results in the highest \ac{hmean} for all three models, suggesting that multi-perspective training increases model robustness and generalisation for text detection. Fine-tuning on the combined dataset and testing on either the aerial or ground perspective results in improved performance compared to fine-tuning solely on the respective perspective's dataset. This indicates that both aerial and ground perspectives benefit from each other during the fine-tuning process. This is not observed in instance segmentation, since \acp{tu} are 3D objects that can appear visually different from different perspectives, while text can be considered as a 2D object that appears similar across perspectives.

\vspace{2mm}
\noindent\textbf{Text Recognition:} For the final stage, we fine-tuned three text recognition models, which were initially pre-trained on the ICDAR15 dataset~\cite{karatzas2015icdar}: SVTR~\cite{du2205svtr}, a model that uses transformer encoder-decoder architecture, RobustScanner~\cite{yue2020robustscanner}, a model that employs a feature fusion model and suitable for contextless text recognition like ISO6346 compliant ID code, and SAR~\cite{pei2017sar}, an early model using a 2D attention mechanism. During the fine-tuning, the model's input were individually cropped text areas based on the ground truth annotations.

\begin{table}
    \centering
    \caption{Results for stage 3: text recognition. Since the ID codes are always uppercase and do not include symbols, the case and symbols are ignored. "-C" indicates that the model is fine-tuned on the Combined dataset but tested on either aerial or ground perspective.}
    {\footnotesize
    \begin{tabular}{@{}llccc@{}}
        \toprule
        \makecell{Perspective} & \makecell{Model} & \makecell{Word acc.} & \makecell{Char. recall} & \makecell{Char. precision} \\
        \midrule
        \multirow{4}{*}{Aerial} & SVTR & 0.60 & 0.86 & \textbf{0.90} \\
        & RobustScanner & 0.64 & 0.87 & 0.87 \\
        & RobustScanner-C & \textbf{0.68} & \textbf{0.88} & 0.88 \\
        & SAR & 0.10 & 0.18 & 0.36 \\
        \cmidrule(lr){1-5}
        \multirow{4}{*}{Ground} & SVTR & 0.49 & 0.69 & \textbf{0.82} \\
        & RobustScanner & 0.50 & 0.70 & 0.72 \\
        & RobustScanner-C & \textbf{0.54} & \textbf{0.73} & 0.73 \\
        & SAR & 0.05 & 0.21 & 0.33 \\
        \cmidrule(lr){1-5}
        \multirow{3}{*}{Combined} & SVTR & 0.57 & 0.81 & \textbf{0.90} \\
        & RobustScanner & \textbf{0.63} & \textbf{0.83} & 0.84 \\
        & SAR & 0.13 & 0.39 & 0.43 \\
        \bottomrule
    \end{tabular}
    }
    \label{tab:text_recognition_results}
\end{table}



\cref{tab:text_recognition_results} shows the results of the three text recognition models that have been used for automatic container code recognition. These models are evaluated on character-level performance (recall and precision) and word-level accuracy. In this context, a word could be the whole ISO6346 compliant ID code, size and type code, or for some trailers the registration plate number written on the \acp{tu}. RobustScanner achieves the best balance of high word accuracy and character-level performance across all perspectives. SVTR has a similar though slightly lower performance compared to RobustScanner in word accuracy. SAR, on the other hand, performs significantly worse than the other two, indicating difficulty to recognise container or trailer ID codes regardless of the perspective. Similar to the text detection stage, fine-tuning on the combined dataset results in better performance in all perspectives.

Compared to the aerial perspective, the three text recognition models underperform when dealing with the ground perspective. Unlike \acp{uav}, ground vehicles do not stop to capture images, resulting in increased motion blur which hinders the performance despite the potentially shorter distance between the camera and the text. Additionally, vertical codes, common on the sides of \acp{tu}, are more prominent in the ground perspective. Occlusion is also more common than in aerial images due to the presence of other objects or people. Because of these reasons, the ground perspective is the most challenging perspective in \ac{trudi}.

\begin{table}
    \centering
    \caption{End-to-end evaluation results of \ac{titus} across aerial, ground, and combined perspectives, showing precision, recall, F1 score, and accuracy.}
    {\footnotesize
    \begin{tabular}{@{}lcccc@{}}
        \toprule
        Perspective & Precision & Recall & F1 Score & Accuracy \\
        \midrule
        Aerial      & \textbf{0.45} & \textbf{0.30} & \textbf{0.36} & \textbf{0.22} \\
        Ground      & 0.25 & 0.19 & 0.22 & 0.12 \\
        Combined    & 0.39 & 0.27 & 0.32 & 0.19 \\
        \bottomrule
    \end{tabular}
    }
    \label{tab:end_to_end_results}
\end{table} 

\vspace{2mm}
\noindent\textbf{End-to-End Evaluation:} Using the best performing models on the combined set, we evaluated \ac{titus} to assess its end-to-end identification performance. The results in \cref{tab:end_to_end_results} highlight benchmarking capabilities of \ac{trudi} and the challenges it offers. Despite the complexity of real-world data, the proposed pipeline is capable of effectively identifying \acp{tu}. In practical deployments, mobile cameras are expected to capture video streams rather than static images, offering temporal continuity that can significantly enhance identification. This continuous input give the system with multiple opportunities to observe a \ac{tu} in consecutive frames, thereby increasing the likelihood of correct recognition and association, even under challenging conditions.

\section{Conclusion} \vspace{-0.5em}
The \ac{trudi} dataset, comprising 35,034 instances of five classes, addresses the need for a publicly available benchmark dataset for the \ac{tu} identification task. It encompasses multiple perspectives and real-world operational conditions, including images captured under various lighting and weather conditions. This allows for a more comprehensive evaluation and ensures that models trained on \ac{trudi} can handle real-world scenarios effectively. The proposed \ac{titus} pipelines follows a three-stage approach including 1) TU segmentation, 2) ID text detection, and 3) text recognition. It offers a robust and flexible solution for TU identification, particularly suitable for mobile cameras mounted on aerial or ground vehicles. The evaluation of the pipeline on the \ac{trudi} dataset demonstrates its effectiveness for \ac{tu} identification. These results set a strong baseline for future research for each individual processing step and the entire \ac{tu} identification pipeline.
The contributions of \ac{trudi} and \ac{titus} are expected to facilitate the development of new applications and methods in TU identification, enhance benchmarking, and improve operational efficiency in multipurpose port logistics.


\section*{Acknowledgements}
The project is supported by the German Federal Ministry for Digital and Transport (BMDV) in the funding program Innovative Hafentechnologien II (IHATEC II). 


\bibliography{egbib}
\end{document}